# Git Loss for Deep Face Recognition

Alessandro Calefati[*1]
a.calefati@uninsubria.it

Muhammad Kamran Janjua[*2]
mjanjua.bscs16seecs@seecs.edu.pk

Shah Nawaz[*1]
snawaz@uninsubria.it

Ignazio Gallo[1]
ignazio.gallo@uninsubria.it

[1] University of Insubria.
Department of Theoretical and Applied Science.
Varese, Italy

[2] National University of Sciences and Technology.
School of Electrical Engineering and Computer Science.
Islamabad, Pakistan

## Abstract

Convolutional Neural Networks (CNNs) have been widely used in computer vision tasks, such as face recognition and verification, and have achieved state-of-the-art results due to their ability to capture discriminative deep features. Conventionally, CNNs have been trained with softmax as supervision signal to penalize the classification loss. In order to further enhance discriminative capability of deep features, we introduce a joint supervision signal, Git loss, which leverages on softmax and center loss functions. The aim of our loss function is to minimize the intra-class variations as well as maximize the inter-class distances. Such minimization and maximization of deep features is considered ideal for face recognition task. We perform experiments on two popular face recognition benchmarks datasets and show that our proposed loss function achieves maximum separability between deep face features of different identities and achieves state-of-the-art accuracy on two major face recognition benchmark datasets: Labeled Faces in the Wild (LFW) and YouTube Faces (YTF). However, it should be noted that the major objective of Git loss is to achieve maximum separability between deep features of divergent identities. The code has also been made publicly available[1].

## 1 Introduction

The current decade is characterized by the widespread use of deep neural networks for different tasks [18, 25, 30]. Similarly, deep convolutional networks have brought about a revolution in face verification, clustering and recognition tasks [6, 18, 19, 27, 29]. Majority of face recognition methods based on deep convolutional networks (CNNs) differ along three primary attributes as explained in [6, 7]. The first is the availability of large scale datasets for training deep neural networks. Datasets such as VGGFace2 [4], CASIA-WebFace [31], UMDFaces [2], MegaFace [13] and MS-Celeb-1M [8] contain images ranging from thousands to millions. The second is the emergence of powerful and scalable network architectures such as Inception-ResNet [10] to train on large scale datasets. The last attribute is the

---

*Equal contribution of these authors to this work.
[1]See code at https://github.com/kjanjua26/Git-Loss-For-Deep-Face-Recognition

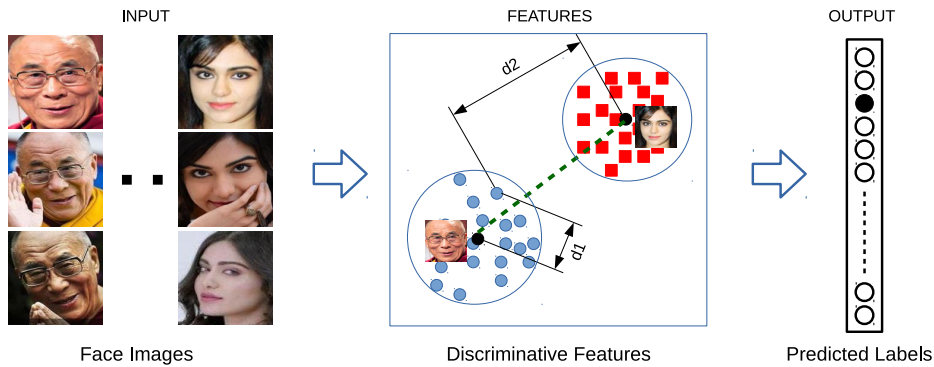

INPUT                              FEATURES                              OUTPUT

Face Images                   Discriminative Features                  Predicted Labels

Figure 1: A toy example depicting the aim of our work: a CNN trained for face recognition supervised by our Git loss function that maximizes the distance $d2$ between features and centroids of different classes and minimizes the distance $d1$ between features and the centroid of the same class.

development of loss functions to effectively modify inter and intra-class variations such as Contrastive loss [20], Triplet loss [19] and Center loss [27], given that softmax penalizes only the overall classification loss.

In this paper, we employ all three attributes associated with face recognition. We use a large scale publicly available dataset, VGGFace2, to train the powerful Inception ResNet-V1 network. We propose a new loss function named Git loss to enhance the discriminative power of deeply learned face features. Specifically, the Git loss simultaneously minimizes intra-class variations and maximizes inter-class distances. A toy example that explains our approach is shown in Figure 1. The name of the loss function is inspired from two common Git version control software commands, "push" and "pull", which are semantically similar to the aim of this work: push away features of different identities while pulling together features belonging to the same identity.

In summary, main contributions of our paper include:

– A novel loss function which leverages on softmax and center loss to provide segregative abilities to deep architectures and enhance the discrimination of deep features to further improve the face recognition task

– Easy implementation of the proposed loss function with standard CNN architectures. Our network is end-to-end trainable and can be directly optimized by fairly standard optimizers such as Stochastic Gradient Descent (SGD).

– We validate our ideas and compare Git loss against different supervision signals. We evaluate the proposed loss function on available datasets, and demonstrate state-of-the-art results.

The organization of the paper is as follows: we review the literature on face recognition in Section 2 and introduce our supervision signal with details in Section 3. We discuss experimental results in Section 4 followed by conclusion and future work in Section 5 and 6.

## 2    Related Work

Recent face recognition works are roughly divided into four major categories: (i) Deep metric learning methods, (ii) Angle-based loss functions, (iii) Imbalanced classes-aware loss functions and (iv) Joint supervision with Softmax. These methods have the aim of enhancing the discriminative power of the deeply learned face features. Deep learning methods [16, 19, 20] successfully employed triplet and contrastive loss functions for face recognition tasks. However, space and time complexities are higher due to the exponential growth of the datasets cardinality.

## 2.1 Deep Metric Learning Approaches

Deep metric learning methods focus on optimizing the similarity (contrastive loss [5, 9]) or relative similarity (triplet loss [11, 26]) of image pairs, while contrastive and triplet loss effectively enhance the discriminative power of deeply learned face features, we argue that both these methods can not constrain on each individual sample and require carefully designed pair and/or triplets. Thus, they suffer from dramatic data expansion while creating sample pairs and triplets from the training set with space complexity being $\mathcal{O}(n^3)$ for triplet networks..

## 2.2 Angle-based Loss Functions

Angular loss constrains the angle at the negative point of triplet triangles, leading to an angle and scale invariant method. In addition, this method is robust against the large variation of feature map in the dataset. ArcFace [7] maximizes decision boundary in angular space based on the L2 normalized weights and features.

## 2.3 Class Imbalance-Aware Loss Functions

Majority of loss functions do not penalize long tail distributions or imbalanced datasets. Range loss [33] employs the data points occurring in the long tail during the training process to get the $k$ greatest ranges harmonic mean values in a class and the shortest inter-class distance in the batch. Although range loss effectively reduces kurtosis of the distribution, it requires intensive computation, hampering the convergence of the model. Furthermore, inter-class maximization is limited because a mini-batch contains only four identities. Similarly, center-invariant loss [29] handles imbalanced classes by selecting the center for each class to be representative, enforcing the model to treat each class equally regardless to the number of samples.

## 2.4 Joint Supervision with Softmax

In joint supervision with softmax based methods [21, 27], the discriminative power of the deeply learned face features is enhanced. The work in [27] penalizes the distance between deep features and their corresponding centers to enhance the discriminative ability of the deeply learned face features. With joint supervision of softmax loss and center loss function, inter-class dispersion and intra-class compactness is obtained. However, this comes with the cost of drastic memory consumption with the increase of CNN layers. Similarly, marginal loss [6] improves the discriminative ability of deep features by simultaneously minimizing the intra-class variances as well as maximizing the inter-class distances by focusing on marginal samples.

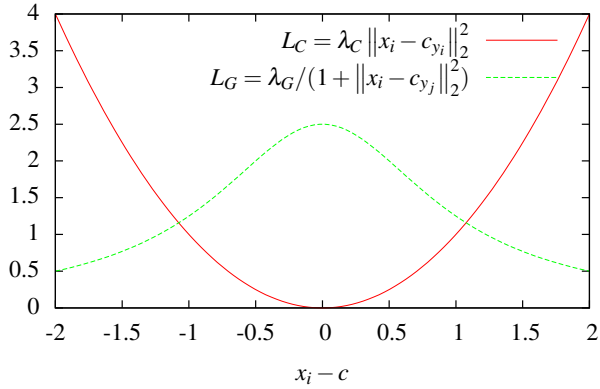

Figure 2: Graphical representation of $L_C$ and $L_G$ varying the distance $(x_i - c)$ in the range $[-2, 2]$. The $L_G$ function takes a maximum value of $\lambda_G$ at $x_i - c = 0$ and has an horizontal asymptote $L_G = 0$.

Inspired from two available works in the literature [6, 27], we propose a new loss function with joint supervision of softmax to simultaneously minimize the intra-class variations and maximize inter-class distances.

## 3  The Git Loss

In this paper, we propose a new loss function called Git loss inspired from the center loss function proposed in [27]. The center loss combines the minimization of the distance between the features of a class and their centroid with the softmax loss to improve the discriminating power of CNNs in face recognition.

In this work, to further improve the center loss function, we add a novel function that maximizes the distance between deeply learned features belonging to different classes (push) while keeping features of the same class compact (pull). The new Git loss function is described in Equation 1:

$$
\begin{aligned}
L &= L_S + \lambda_C L_C + \lambda_G L_G \\
&= -\sum_{i=1}^{m} log \frac{e^{W_{y_i}^T x_i + b_{y_i}}}{\sum_{j=1}^{n} e^{W_j^T x_i + b_j}} + \frac{\lambda_C}{2} \sum_{i=1}^{n} \|x_i - c_{y_i}\|_2^2 + \lambda_G \sum_{i,j=1, i \neq j}^{m} \frac{1}{1 + \|x_i - c_{y_j}\|_2^2}
\end{aligned}
\tag{1}
$$

where $L_G$ is equal to $\frac{1}{1+\|x_i-c_{y_j}\|_2^2}$ which is responsible for maximizing the distance between divergent identities. The deep features of the $i$-th samples belonging to the $y_i$-th identity are denoted by $x_i \in \mathbb{R}^d$. The feature dimension $d$ is set as 128, as reported in [19]. $W_j \in \mathbb{R}^d$ denotes the $j$-th column of the weights $W \in \mathbb{R}^{d \times n}$ in the last fully connected layer and $b \in \mathbb{R}^n$ is the bias term. $c_{y_i}$ is the center of all deep features $x_i$ belonging to the $y_i$-th identity. When the parameter $\lambda_G = 0$ the center loss function can be obtained.

The gradient $(\frac{\partial L_G}{\partial x_i})$ of the $L_G$ with respect to $x_i$ can be computed as:

$$
\frac{\partial L_G}{\partial x_i} = \frac{\partial}{\partial x_i} \left( \frac{1}{1 + \|x_i - c_{y_j}\|_2^2} \right)
\tag{2}
$$

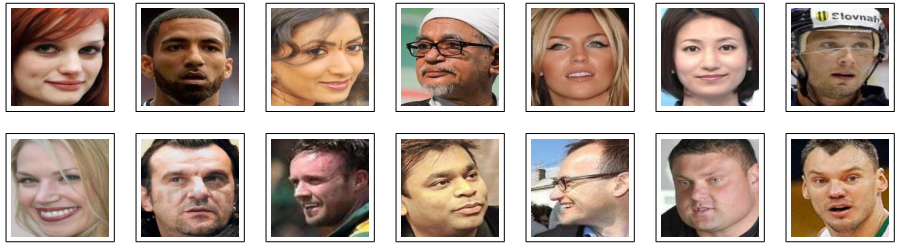

Figure 3: Some sample images taken from the VGGFace2 dataset aligned and cropped to $160 \times 160$ pixels.

Let $u = 1 + \|x_i - c_{y_j}\|_2^2$, thus $f = u^{-1}$. The equation 2 can be solved to compute the final gradient. We substitute the values of $u$ and get $\frac{\partial u^{-1}}{\partial u} \frac{\partial}{\partial x_i}(1 + \|x_i - c_{y_j}\|_2^2)$. Solving $\frac{\partial u^{-1}}{\partial u}$ and $\frac{\partial}{\partial x_i}(1 + \|x_i - c_{y_j}\|_2^2)$, we get $\frac{-1}{u^2}$ and $2(x_i - c_{y_j})$ respectively. Substituting these values, the final equation becomes $\frac{-1}{u^2}(2(x_i - c_{y_j}))$. We can simplify this equation to obtain the final gradient equation 3.

$$= \frac{-2(x_i - c_{y_j})}{(1 + (x_i - c_{y_j})^2)^2} \tag{3}$$

Git loss simultaneously minimizes the intra-class variances using the $L_C$ function and maximizes the inter-class distances using the $L_G$ function. Parameters $\lambda_C$ and $\lambda_G$ are used to balance the two functions $L_C$ and $L_G$ respectively. From the plot of $L_C$ and $L_G$ functions, shown in Figure 2, it can be observed how these two functions have opposite behaviors: to minimize $L_C$ we have to reduce the distance between features and the centers, while to maximize $L_G$ we must maximize the distance between features and all centroids of other classes. $L_G$ is a continuous and differentiable function, thus it can be used to train CNNs optimized by the standard Stochastic Gradient Descent (SGD) [15].

# 4 Experiments

We report experimental results on currently popular face recognition benchmark datasets, Labeled Faces in the Wild (LFW) [12] and YouTube Faces (YTF) [28]. We also report a set of experiments to investigate the hyper-parameters associated with the loss function.

## 4.1 Experimental Settings

### 4.1.1 Training Data.

We use data from VGGFace2 [4] dataset to train our model. It contains 3.31 million images of 9,131 identities, with an average of 362.6 images for each identity. Moreover, the dataset is characterized by a large range of poses, face alignments, ages and ethnicities. The dataset is split into train and test sets with 8,631 and 500 identities respectively, but we only used the train set. Some representative images taken from the VGGFace2 dataset are shown in Figure 3.

### 4.1.2 Data Preprocessing.

The label noise is minimized through automated and manual filtering in the VGGFace2 dataset. We applied horizontal flipping and random cropping data augmentation techniques

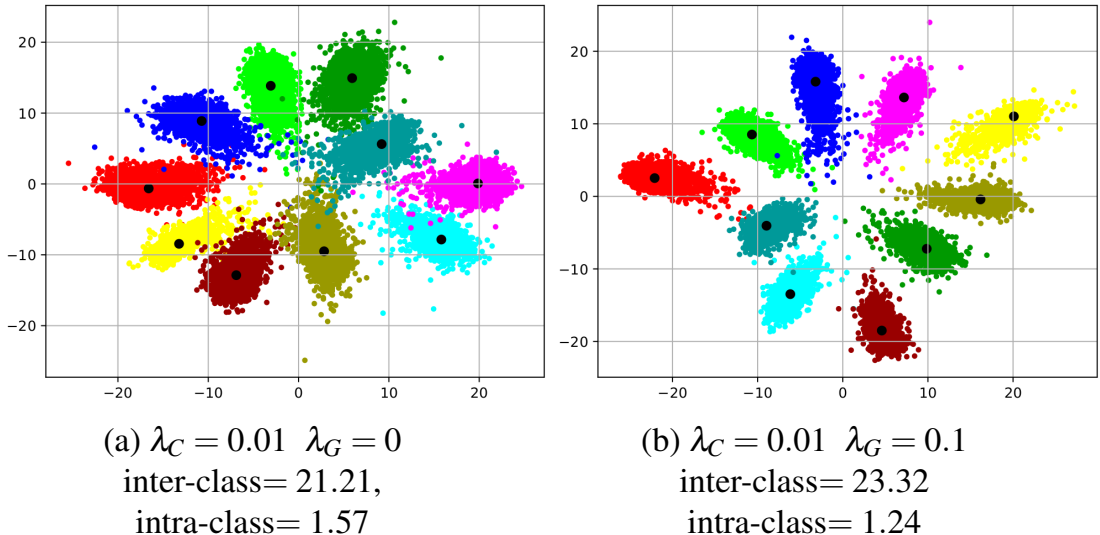

(a) $\lambda_C = 0.01$   $\lambda_G = 0$
inter-class= 21.21,
intra-class= 1.57

(b) $\lambda_C = 0.01$   $\lambda_G = 0.1$
inter-class= 23.32
intra-class= 1.24

Figure 4: Two plots showing the behavior of Center loss (a) and Git loss (b) on MNIST training set. Using the Git loss function features are more compact (smaller intra-class distances) and more spaced (larger inter-class distances), further enhancing the discriminative power of deep features. Points with different colors denote features from different classes. (best viewed in color)

to images, then face images are aligned using the Multi-Task CNN [32] and finally cropped to a size of $160 \times 160$ pixels before feeding to the network. We noticed that VGGFace2 contains 496 overlapping identities with LFW and 205 with YTF datasets, therefore, we removed overlapping identities from both datasets to report fair results.

### 4.1.3   Network Settings.

We implemented the proposed Git loss in Tensorflow [1] and the network was trained using Nvidia's GeForce GTX 1080 GPU. The implementation is inspired from the *facenet* work, available on Github[2]. We employ the Inception ResNet-V1 network architecture and process 90 images in a batch. We use adaptive learning rate for the training process with a starting value of $-1$ and decreased it by a factor of 10 with Adam Optimizer [14], thus adding robustness to noisy gradient information and various data modalities across the dataset, improving the performance of the final model.

### 4.1.4   Test Settings.

Deep face features ($128 - dimensional$) are taken from the output of the fully connected layer. Since the features are projected to Euclidean space, the score is computed using the Euclidean distance between two deep features. Threshold comparison is obtained with $10 - fold$ cross validation for verification task. We employ two different trained models for LFW and YTF datasets due to number of overlapping identities with VGGFace2 dataset.

Table 1: Comparison between Center loss ($\lambda_C$) and Git loss ($\lambda_G$) on MNIST dataset. Values are obtained by averaging 10 runs. We highlighted best results compared to Center loss ($\lambda_G = 0$) for each configuration. We achieved reduced intra-class distance along with higher inter-class distance compared to Center loss.

| $\lambda_C$ | $\lambda_G$ | Loss | Train Acc.(%) | Val. Acc.(%) | Inter Dist. | Intra Dist. |
|---|---|---|---|---|---|---|
| 0.0001 | 0 | 0.020 | 99.77 | 98.52 | 85.95 | 8.39 |
| 0.0001 | 0.0001 | 0.0132 | 100.00 | 98.65 | 87.87 | 8.52 |
| **0.0001** | **0.001** | **0.016** | **99.77** | **98.66** | **89.82** | **8.20** |
| 0.0001 | 0.01 | 0.020 | 99.77 | 98.62 | 88.48 | 8.54 |
| 0.0001 | 0.1 | 0.032 | 99.61 | 98.46 | 96.76 | 9.56 |
| 0.0001 | 1 | 0.466 | 89.77 | 88.95 | 137.37 | 15.45 |
| 0.0001 | 1.5 | 0.641 | 80.63 | 79.56 | 160.28 | 16.91 |
| 0.0001 | 2 | 1.001 | 69.84 | 68.51 | 125.84 | 17.78 |
| 0.001 | 0 | 0.021 | 99.61 | 98.67 | 44.36 | 3.10 |
| 0.001 | 0.0001 | 0.117 | 96.75 | 95.66 | 51.77 | 4.75 |
| 0.001 | 0.001 | 0.024 | 99.84 | 98.53 | 47.81 | 3.23 |
| **0.001** | **0.01** | **0.025** | **99.77** | **98.62** | **46.13** | **3.16** |
| 0.001 | 0.1 | 0.053 | 99.22 | 98.69 | 51.22 | 3.46 |
| 0.001 | 1 | 0.779 | 76.10 | 76.33 | 68.77 | 6.55 |
| 0.001 | 1.5 | 0.460 | 89.22 | 89.06 | 89.67 | 7.45 |
| 0.001 | 2 | 0.757 | 80.94 | 81.89 | 96.02 | 9.14 |
| 0.01 | 0 | 0.025 | 99.65 | 98.89 | 21.36 | 1.09 |
| 0.01 | 0.0001 | 0.031 | 99.69 | 98.77 | 22.07 | 1.16 |
| 0.01 | 0.001 | 0.024 | 100.00 | 98.75 | 20.63 | 1.18 |
| 0.01 | 0.01 | 0.051 | 99.53 | 98.61 | 21.96 | 1.09 |
| **0.01** | **0.1** | **0.037** | **99.84** | **98.70** | **22.93** | **1.22** |
| 0.01 | 1 | 0.937 | 71.17 | 71.38 | 30.25 | 2.02 |
| 0.01 | 1.5 | 0.368 | 97.58 | 96.50 | 50.56 | 3.10 |
| 0.01 | 2 | 0.824 | 83.44 | 84.10 | 46.35 | 3.03 |
| 0.1 | 0 | 0.040 | 99.74 | 98.89 | 9.76 | 0.38 |
| 0.1 | 0.0001 | 0.049 | 99.53 | 98.85 | 9.65 | 0.42 |
| 0.1 | 0.001 | 0.024 | 100.00 | 98.96 | 10.33 | 0.38 |
| 0.1 | 0.01 | 0.026 | 99.92 | 99.00 | 9.76 | 0.37 |
| **0.1** | **0.1** | **0.040** | **100.00** | **98.96** | **10.99** | **0.37** |
| 0.1 | 1 | 1.508 | 57.11 | 57.90 | 10.52 | 0.55 |
| 0.1 | 1.5 | 1.741 | 53.59 | 54.03 | 10.81 | 1.03 |
| 0.1 | 2 | 1.536 | 67.98 | 66.65 | 15.43 | 1.03 |
| 1 | 0 | 0.031 | 100.00 | 99.00 | 5.12 | 0.14 |
| 1 | 0.0001 | 0.178 | 96.72 | 95.72 | 5.86 | 0.22 |
| 1 | 0.001 | 0.023 | 100.00 | 99.03 | 4.94 | 0.12 |
| 1 | 0.01 | 0.027 | 100.00 | 99.04 | 5.03 | 0.12 |
| 1 | 0.1 | 0.064 | 99.92 | 99.04 | 5.25 | 0.14 |
| **1** | **1** | **0.264** | **99.92** | **99.02** | **8.30** | **0.21** |
| 1 | 1.5 | 0.330 | 100.00 | 98.96 | 9.73 | 0.23 |
| 1 | 2 | 0.847 | 91.10 | 89.79 | 9.22 | 0.25 |

## 4.2  Experiments with $\lambda_C$ and $\lambda_G$ parameters

Parameters $\lambda_C$ and $\lambda_G$ are used to balance two loss functions $L_C$ and $L_G$ with softmax. In our model, $\lambda_C$ controls intra-class variance while $\lambda_G$ controls inter-class distances. We conducted various experiments to investigate the sensitiveness of these two parameters. These tests are systematic random search heuristics based. The major reason for employing heuristic methodologies over techniques like GridSearch is that when the dimensionality is high, the number of combinations to search becomes enormous and thus techniques like Grid-Search become an overhead. The work in [3] argues why performance of GridSearch is not satisfactory as compared to other techniques. Table 1 shows average result values over 10 runs on MNIST dataset, we have following outcomes: (i) Smaller values of $\lambda_C$ increase inter-class distance, but they also increase intra-class distances which is undesirable in face recognition. (ii) Our loss function produces higher inter-class distances and lower intra-class distances. An example displaying the qualitative and quantitative results of our Git loss and Center loss is shown in Figure 4. Note that these results are obtained with a single run on MNIST dataset.

## 4.3  Experiments on LFW and YTF datasets

We evaluate the proposed loss function on two famous face recognition benchmark datasets: LFW [12] and YTF [28] in unconstrained environments i.e. under open-set protocol. LFW dataset contains 13,233 web-collected images from 5,749 different identities, with large variations in pose, expression and illumination. We follow the standard protocol of *unrestricted with labeled outside data* and tested on 6,000 face pairs. Results are shown in Table 2.

YTF dataset consists of 3,425 videos of 1,595 different people, with an average of 2.15 videos per person. The duration of each video varies from 48 to 6,070 frames, with an average length of 181.3 frames. We follow the same protocol and reported results on 5,000 video pairs in Table 2.

We compare the Git loss against many existing state-of-the-art face recognition methods in Table 2. From results, we can see that the proposed Git loss outperforms the softmax loss by a significant margin, from 98.40% to 99.30% in LFW and from 93.60% to 95.30% in YTF. In addition, we compare our results with center loss method using the same network architecture (Inception-ResNet V1) and dataset (VGGFace2). The Git loss outperforms the center loss, obtaining an accuracy of 99.30% as compared to 99.20% on LFW and 95.30% compared to 95.10% on YTF. These results indicate that the proposed Git loss further enhances the discriminative power of deeply learned face features. Moreover, we trained our model with $\approx$ 3M images which are far less than other state-of-the-art methods such as [19, 23, 24], reported in Table 2.

## 5  Concluding Remarks

This paper proposes a new loss function, named Git loss, which makes deep feature more discriminable. We exploit the softmax as supervision signal and the well-known property of the center loss which compacts patterns of a class, lowering intra-class distances. The result is a new loss function which minimizes intra-class variations and maximizes inter-class distances simultaneously. We evaluate capabilities of the new loss function on two common face benchmark datasets such as LFW and YTF. Results obtained clearly demonstrate the effectiveness and generalization abilities of the proposed loss function. However, the major objective of Git loss is achieving maximum separability between dissimilar classes and compactness between similar classes.

| Methods | Images | LFW(%) | YTF(%) |
|---------|--------|--------|--------|
| DeepID [20] | - | 99.47 | 93.20 |
| VGG Face [18] | 2.6M | 98.95 | 97.30 |
| Deep Face [23] | 4M | 98.37 | 91.40 |
| Fusion [24] | 500M | 98.37 | - |
| FaceNet [19] | 200M | 99.63 | 95.10 |
| Baidu [16] | 1.3M | 99.13 | - |
| Range Loss [34] | 1.5M | 99.52 | 93.70 |
| Multibatch [22] | 2.6M | 98.80 | - |
| Aug [17] | 0.5M | 98.06 | - |
| Center Loss [27] | 0.7M | 99.28 | 94.90 |
| Marginal Loss [6] | 4M | 99.48 | 95.98 |
| Softmax | $\approx$ 3M | 98.40 | 93.60 |
| Center Loss [27] | $\approx$ 3M | 99.20 | 95.10 |
| Git Loss (Ours) | $\approx$ 3M | **99.30** | **95.30** |

Table 2: Performance verification of different state-of-the-art methods on LFW and YTF datasets. The last three rows show results using the same architecture (Inception-ResNet V1) trained on the VGGFace2 dataset.

## 6 Future Work

In future, we would like to extend Git loss to deal with class imbalance problem where penalization of data points occurring in long tail is not effective. We would also like to explore different baseline architectures and greater number of baseline images to further the experimental analysis. The objective of Git loss is not just empirical results alone, but the achieving of discriminatory ability in feature space. Furthermore, we believe that Git loss can be extended to multiple modalities where data from different input streams is combined and projected on a common feature space and thus would like to explore this perspective as well.